\documentclass{elsarticle}

\usepackage{lineno,hyperref}
\usepackage{times}

\usepackage{graphicx}
\usepackage{amsmath}
\usepackage{amssymb}
\usepackage{multirow}
\usepackage{color}
\usepackage{caption}
\usepackage{subcaption}
\modulolinenumbers[5]

\journal{and Accepted by ISPRS Journal of Photogrammetry and Remote Sensing}

\begin{document}

\begin{frontmatter}

\title{PoseGAN: A Pose-to-Image Translation Framework for Camera Localization}

\author[mymainaddress,secondaddress,thirdaddress]{Kanglin Liu}
\author[mymainaddress,fourthaddress]{Qing Li}

\author[mymainaddress,secondaddress,thirdaddress,fourthaddress]{Guoping Qiu \corref{mycorrespondingauthor}}
\cortext[mycorrespondingauthor]{Corresponding author}
\ead{guoping.qiu@nottingham.ac.uk}

\address[mymainaddress]{Shenzhen University, Shenzhen, China}
\address[secondaddress]{Guangdong Key Laboratory of Intelligent Information Processing, Shenzhen, China}
\address[thirdaddress]{Shenzhen Institute of Artificial Intelligence and Robotics for Society, Shenzhen, China}
\address[fourthaddress]{University of Nottingham, Nottingham, United Kingdom.}

\begin{abstract}
Camera localization is a fundamental requirement in robotics and computer vision.
This paper introduces a pose-to-image translation framework to tackle the camera localization problem. We present PoseGANs, a conditional generative adversarial networks (cGANs) based framework for the implementation of pose-to-image translation. PoseGANs feature a number of innovations including a distance metric based conditional discriminator to conduct camera localization and a pose estimation technique for generated camera images as a stronger constraint to improve camera localization performance.  Compared with learning-based regression methods such as PoseNet, PoseGANs can achieve better performance with model sizes that are 70\% smaller. In addition, PoseGANs introduce the view synthesis technique to establish the correspondence between the 2D images and the scene, \textit{i.e.}, given a pose, PoseGANs are able to synthesize its corresponding camera images.
Furthermore, we demonstrate that PoseGANs differ in principle from structure-based localization and learning-based regressions for camera localization, and show that PoseGANs exploit the geometric structures to accomplish the camera localization task, and is therefore more stable than and superior to learning-based regressions which rely on local texture features instead.
In addition to camera localization and view synthesis, we also demonstrate that PoseGANs can be successfully used for other interesting applications such as moving object elimination and frame interpolation in video sequences. The accompanying videos are available at \url{https://drive.google.com/drive/folders/1X--A68s1hctb_QwywhbHpHnIoNpcAXFc?usp=sharing}.

\end{abstract}


\end{frontmatter}


\section{Introduction}
Inferring the camera's absolute pose, or camera localization is a key component of many computer vision tasks like structure from motion (SfM), simultaneous localization and mapping (SLAM) \cite{Engel2014834, MurArtal20151147, Zhou20151364} as well as many applications such as robotics and autonomous driving \cite{Hane201714,Castle200815}. 
Conventional camera localization algorithm  is conducted based on three schemes: structure-based localization \cite{Brachmann20176684,Brachmann20184654,Cavallari20174457,Meng20176886, Sattler20161744}, image retrieval \cite{Radwan20184407} and learning-based regression \cite{Kendall20152938,Melekhov2017879,Kendall20164762,Kendall20175974,Brahmbhatt20182616,Radwan20184407,Ummenhofer20175038}.
Structure-based localization solves this problem by establishing correspondence between pixels in a 2D image and 3D points in the scene via descriptor matching. Such a set of 2D-3D matches allows to estimate camera poses by applying a n-point-pose (PnP) \cite{Albl20152292} solver inside a RANSAC framework \cite{Chum20081472,Fischler1981381}. However, 2D-3D matches are not feasible for many indoor scenes as they are intensively reliance on the feature detector and descriptor. Motion blur, strong illumination, texture-less or repetitive indoor surfaces may result in localization failure.
Image retrieval tries to address the camera localization problem by approximating the pose of a test image by the pose of the most similar retrieved image. More precise estimates can be obtained by using feature matches between the test image and the retrieved images for relative pose estimation. 
Learning-based regressions apply the deep convolutional neural network as the camera pose regressors, either directly estimating the pose through regression \cite{Kendall20152938} or predicting the pose of a test image relative to one or more training images \cite{Laskar2017929}.  Learning-based methods are computationally efficient. Yet, they are also significantly less accurate than structure-based localization, and are found to be closely related to image retrieval in principle \cite{Sattler20193302}.

In this paper, we set out to present a novel insight to the camera localization problem and propose a new framework to solve it.
Given a specific camera and a scene, the camera’s sighted scene is determined by its 6-DoF pose, or equivalently, the content of the image taken by the camera is determined by the camera’s pose.   To be specific, a camera shot image $x \in \mathbb{R}^{H\times W \times 3} $ can be regarded as a sampling from a specific scene, where $H$ and $W$ are the height and width of the image, respectively. Meanwhile, camera pose $y=(p, r)$, a $(3+d)$-dimensional continuous variable,  describes the camera trajectory, where $p \in \mathbb{R}^{3} $ and $r \in \mathbb{R}^{d} $ represent the position and orientation of the camera, respectively. There are multiple ways to represent the orientation $r$, \textit{e.g.}, as a $4$-dimensional  unit quaternion, or a $3$-dimensional vector representing the Euler angle. Obviously,  there are two latent distributions $p_x$ and $p_y$, which can be applied to describe the distributions $x$ and $y$ obey respectively. 
Moreover, $p_x$ and $p_y$ are conditionally linked up by  the intrinsic relationship that each $x$ corresponds to a unique $y$, and that  each $y$ decides a unique $x$.  
Thus, we  treat the camera localization problem as a pose-to-image translation, and  propose a conditional generative adversarial networks (cGANs) based framework, referred to as PoseGAN, to solve it. 
To be specific, the generator $G$ would synthesize the camera shot image $x^{'} = G(y) \in p_{g}$ conditioned on a given pose $y$, where $p_g$ represents the generated distribution, and then the discriminator $D$ would distinguish the real pairs $(x, y)$ from the fake pairs $(G(y), y)$, where pose estimation is included in $D$ to serve as the conditional discrimination. With the adversarial training between $G$ and $D$,  $G$ would output realistic camera shot images, and $D$ would eventually have a good performance on predicting the pose of the given images. 

Owing to the contribution of the vast generated images to supervising the training of $D$, PoseGAN has been demonstrated to able to achieve good performance on camera localization, even with model sizes $70\%$ smaller than that of  previous models like PoseNet. Furthermore, we experimentally demonstrate that PoseGAN exploits geometry structures as evidences to predict the poses, superior to learning-based regressions, which use local texture features instead.
In addition, PoseGAN derives a number of promising applications, \textit{e.g.}, given a pose, the generator $G$ allows PoseGAN to generate its corresponding image, which is referred to as view synthesis or image rendering. The view synthesis technique in PoseGAN establishes the correspondence between the 2D images and the scene, thus can provide more samples for the traditional SfM technique in the 3D scene reconstruction process, and can be also used in some view synthesis based image retrieval frameworks.  Besides, PoseGANs are capble of applications like moving object elimination and frame interpolation.

To summarize, our contributions are as follow:

(1) A novel cGAN based framework for camera localization - PoseGANs. This as far as we know is the first method in the literature that tackles the camera localization problem via a pose-to-image translation. 

(2)  Experiments have demonstrated that PoseGANs achieve good performance on camera localization task, even with model size 70\% smaller than that of previous models like PoseNet.

(3) It has been demonstrated that PoseGAN differ in principle from other camera localization methods, and that PoseGANs exploit the geometry structures to accomplish the task, superior to learning-based regressions.

(4) PoseGANs are capable of appealing applications, \textit{e.g.}, view synthesis, moving object elimination and frame interpolation.  

\section{Related Work}
\subsection{Camera Localization Methods}
Conventional camera localization algorithms conduct pose estimation mainly through three schemes: image retrieval, learning-based regression and structure-based localization.
Learning-based regressions rely on convolutional neural networks (CNNs),  \textit{e.g.}, VGG \cite{Simonyan2014} or ResNet \cite{He2016},  to embeds the images to a high-dimensional space, then extract features for regressing the camera pose.  Since its first introduction by PoseNet \cite{Kendall20152938}, numerous works have been made, mainly focusing on modifying the architecture or loss objectives towards performance promotion \cite{Balntas2018751,Laskar2017929,Melekhov2017675,Saha2018}, \textit{e.g.},  \cite{Kendall20164762,Kendall20175974,Brahmbhatt20182616}  extend PoseNet by using a weighted combination of position and orientation errors.  Visual odometry constraint is used in \cite{Radwan20184407, Ummenhofer20175038} for improving localization precision.
Recently, it has been found that learning-based regressions are inherently more closely related to image retrieval than to structure-based localization.  In contrast to image retrieval and learning-based regression, structure-based localization  predicts the location of the query image in a 3D map through establishing 2D-3D correspondences by matching local features. 
Traditionally, 2D-3D match is based on matching descriptors extracted in the test image against descriptors associated with the 3D points. Alternatively, machine learning techniques can be used to directly regress 3D point positions from image patches \cite{Brachmann20176684,Brachmann20184654}.
Orthogonal to those works, proposed PoseGAN tries to tackle the camera localization problem in a novel way by a pose-to-image translation.

\subsection{Generative Adversarial Networks (GANs)} 
Generative adversarial networks (GANs) \cite{Goodfellow20142672} is a special generative model to learn a generator $G$ to capture the data distribution via an adversarial process. Specifically, a discriminator $D$ is introduced to distinguish the generated images from the real ones, while the generator $G$ is updated to confuse the discriminator. The adversarial process is formulated as a minimax game as \cite{Radford2015}:
\begin{equation} \label{key3-1}
\underset{G}{\mathrm{min}} \ \underset{D}{\mathrm{max}} V(G, D)
\end{equation}
where min and max of $G$ and $D$ are taken over the set of the generator and discriminator functions respectively. $V(G, D)$ is to evaluate the difference in the two distributions of $q_x$ and $q_g$, where $q_x$ is the data distribution, and $q_g$ is the generated distribution.
The conventional form of $V(G, D)$ is given by Kellback-Leibler (KL) divergence: $ E_{x \sim q_{x}}[\mathrm{log}D(x)]+E_{x'\sim q_{g}}[\mathrm{log}(1-D(x'))]$ \cite{Miyato2018,Gulrajani2017,Arjovsky2017}.

Conditional GANs (cGANs) are a type of GANs that use conditional information for the discriminator and generator. Unlike in standard GANs, the discriminator of cGANs discriminates between the generator and the data distribution on the set of the pairs of generated samples and its intended conditional variable \cite{Miyato20182}. Owing to their ability to learn highly structured probability distribution, cGANs have been widely used in applications like  high fidelity image generation \cite{Andrew2018, Karras2019}, image editing \cite{E15} and image-to-image translation \cite{Zhu2017}. In this paper, we introduce PoseGAN, a cGANs based framework, to tackle the camera localization problem based on the motivation that the camera poses can be obtained via a pose-to-image translation.

\section{PoseGAN for Camera Localization} \label{Arch}

\begin{figure}[t]
	\centering
	\begin{subfigure}[b]{0.4\textwidth}
		\centering
		\includegraphics[height=0.5\linewidth]{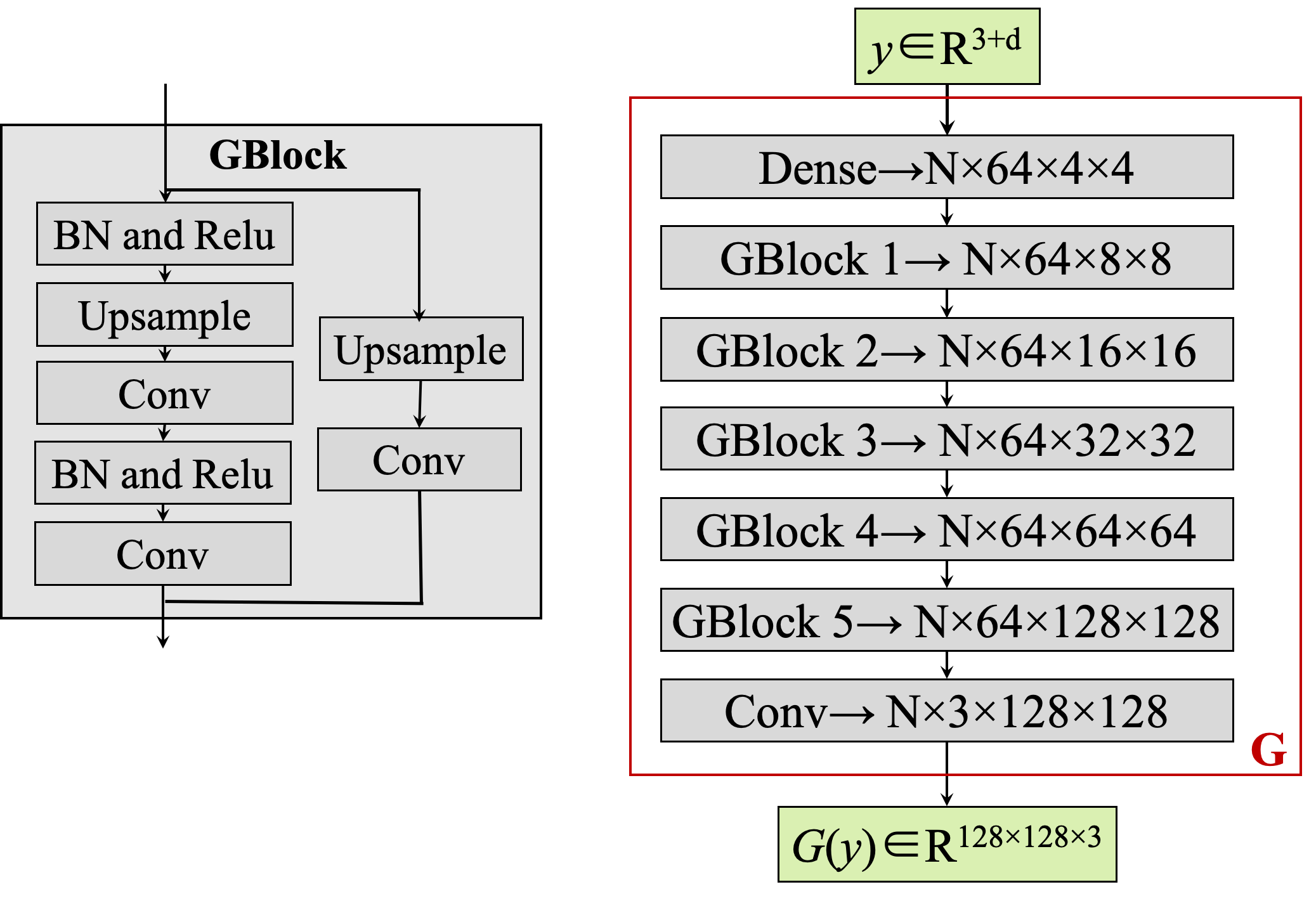}	
		\caption{$G$ architecture}
	\end{subfigure}
	\begin{subfigure}[b]{0.55\textwidth}
		\centering
		\includegraphics[height=0.5\linewidth]{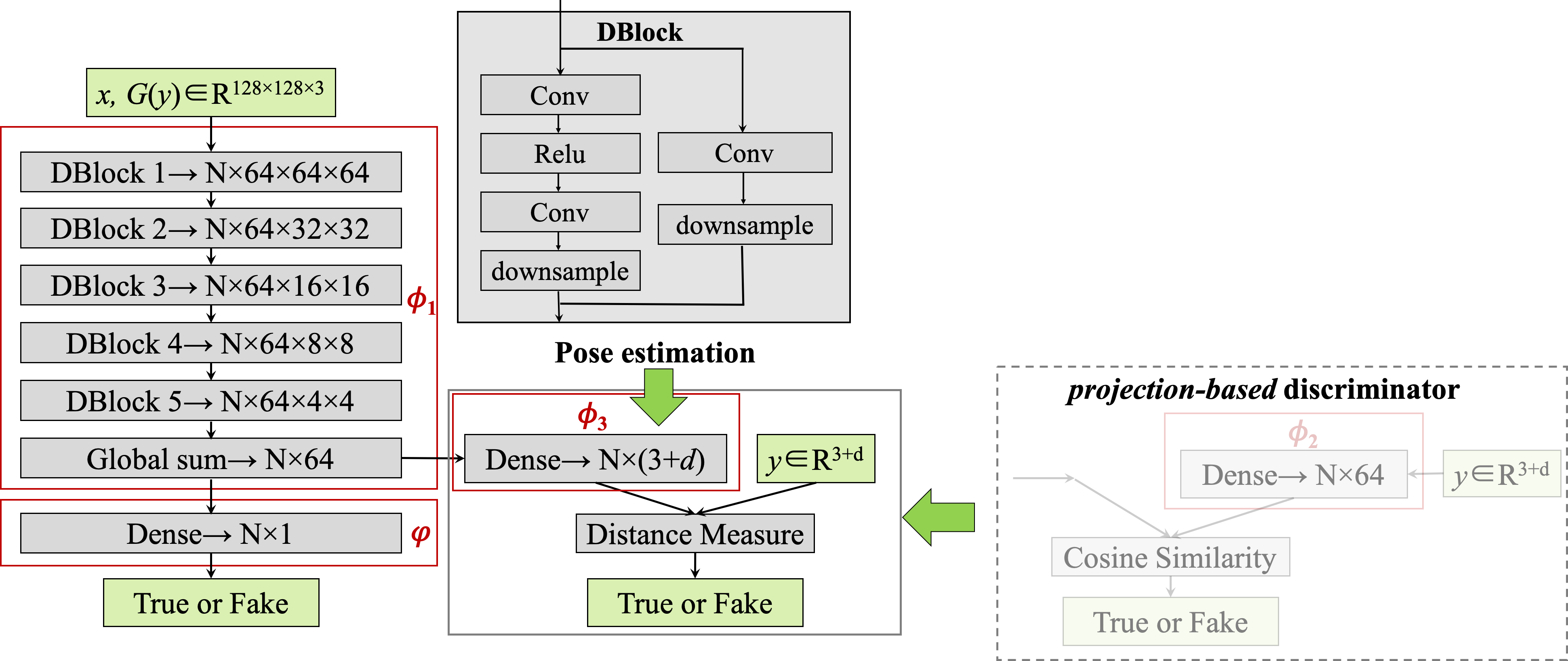}	
		\caption{$D$ architecture}
	\end{subfigure}

	\caption{Illustration of PoseGANs architecture for camera localization. $G$ is mainly comprised of 5 GBlocks, whose details are shown in the left. BN is the batch normalization \cite{Ioffe2015}, Relu is the non-linear operation \cite{Glorot2011}, Conv represents the convolutional operation, and N represents the batch size in the training. $D$ mainly contains 3 parametric functions: $\phi_1(\cdot)$, $\phi_3(\cdot)$ and $\varphi(\cdot)$. Specifically, $\phi_1$ consists of 5 DBlocks, whose details are shown in the right.
	$G$ is responsible for generating camera shot images $G(y)$ conditioned on the given pose $y$, and $D$ is trained to distinguish the true pair$(x,y)$ from the fake pair$(G(y),y)$. 
	The dotted box shows the \textit{projection-based} discriminator which conducts conditional discrimination via cosine similarity, thus incapable of pose estimation.
	PoseGAN conducts distance measure instead, and camera localization is implemented by $\phi_3(\phi_1(\cdot))$.}
	\label{fig:fig1}
	
\end{figure}

This section introduces the PoseGAN for solving the camera localization via a pose-to-image translation. As shown in Figure \ref{fig:fig1}, PoseGAN applies the generator $G$ to generate the camera shot images $G(y)$ at the given pose $y$, where $y$ is treated as the conditional information for sample generations. Then, the discriminator $D$ is trained to distinguish the fake pair $(G(y), y)$ from the real pair $(x, y)$. 
In this paper, we use the \textit{projection-based} discriminator \cite{Miyato20182} as an implementation of $D$ to conduct the conditional discrimination between $(G(y), y)$ and $(x, y)$.  
Conventionally, the outputs of the \textit{projection-based} discriminator are comprised of two parts:
\begin{equation} \label{key3-2}
D(x, y) =\varphi (\phi_1(x))+<\phi_1(x), \phi_2(y)>
\end{equation}
where $\varphi(\cdot)$, $\phi_1(\cdot)$ and $\phi_2(\cdot)$ are some parametric functions as shown in Figure \ref{fig:fig1}, $<\cdot, \cdot>$ determines the cosine similarity. 
The first term $\varphi(\phi_1(x))$ identifies the data distribution. The second term first embeds the image $x$ and pose $y$ to a high-dimensional space via $\phi_1(\cdot)$ and $\phi_2(\cdot)$, respectively, then calculate the cosine similarity between $\phi_1(x)$ and $\phi_2(y)$, thus encouraging the generation of samples whose features match a learned condition prototype.
However,  such a discriminator is incapable of predicting the poses. To solve this, $D$ in PoseGAN conducts distance measures instead of calculating cosine similarity between $\phi_1(x)$ and $\phi_2(y)$.  The outputs of the discriminator is modified as:
\begin{equation}\label{key3-3}
D(x, y) =\varphi (\phi_1(x))+||\phi_3(\phi_1(x))-y||
\end{equation}
where $\phi_3(\cdot)$ is a linear function for projecting the $\phi_1(x)$ to a $(3+d)$-dimensional vector. Then,  $D$ is conditioned by the term $||\phi_3(\phi_1(x))-y||$, where $\phi_3(\phi_1(\cdot))$ allows $D$ to estimate the pose. Similar to $<\phi_1(x), \phi_2(y)>$, $||\phi_3(\phi_1(x))-y||$ can still encourage the generated samples to match the conditional prototype.
The adversarial training between $G$ and $D$ would encourage $G$ to generate realistic camera shot images whose poses correspond to the correct pose, and encourage $\phi_3(\phi_1(\cdot))$ to identify the correct pose of given images for serving as the conditional discrimination. Eventually,  $\phi_3(\phi_1(\cdot))$  would have a good performance on pose prediction.


\section{Loss Functions} \label{loss}

As indicated by Equation \ref{key3-3}, there are two terms in the output of the discriminator.
The purpose of $\varphi(\phi_1 (\cdot))$ is to distinguish the generated data from the real one. 
Given the real data $x$, $\varphi(\phi_1 (x))$ should regard it as the real data. Given the generated data $G(y)$, $\varphi(\phi_1 (G(y)))$ should regard it as the fake data. Applying KL divergence \cite{Goodfellow20142672, Radford2015} and hinge loss \cite{Miyato2018, Miyato20182} to describe the difference between  the real and generated distributions, we can obtain the objective function $\mathcal{L}_1$ for $\varphi(\phi_1 (\cdot))$:
\begin{equation}\label{key4}
\mathcal{L}_1 = E_{x \sim p_{x}}\left [ min(0, -1+\varphi(\phi_1 (x))) \right]+E_{y \sim p_{y}}\left [ min(0, -1-\varphi(\phi_1 (G(y)))) \right]
\end{equation}
where $p_x$, $p_y$ represent the distribution of $x$ and $y$, respectively. $\varphi$ and $\phi_1$ are parametric functions in $D$.

Correspondingly, $D$ is conditioned by $||\phi_3(\phi_1(x))-y||$, which provide evidences for conditional discrimination.  Hence, to indicate the true pair ($x$, $y$), the loss objective should minimize $\left \| \phi_3(\phi_1(x)) - y \right \|$. Intuitively, for generated samples $G(y)$, the loss objective should maximize $\left \| \phi_3(\phi_1(G(y))) - y \right \|$.
However, such an intuition is actually wrong. Because each $y$ in $\mathbb{R}^{3+d}$ has a corresponding $x$ in the pixel domain. Each point away from $y$ actually corresponds to a particular image instead of the generated image $G(y)$. Therefore, it would be wrong to maximize $\left \| \phi_3(\phi_1(G(y))) - y \right \|$ for generated samples $G(y)$.

To solve this, we introduce an estimation method to roughly determine the pose of generated samples $G(y)$.
In PoseGANs, $G(y)$ can be regarded as the estimation of samples in the target distribution with errors. In the same way, the pose of $G(y)$ should be located in a certain range around the pose of $x$.  Because ($G(y)$, $y$) should be regarded as a wrong sample instead of a fake one in the case that $\left \| \phi_3(\phi_1(G(y))) - \phi_3(\phi_1(x)) \right \|$ is large enough. Therefore, the difference between $\phi_3(\phi_1(x))$ and $\phi_3(\phi_1(G(y)))$ should locate in a specific range:

\begin{equation}\label{key5}
\left \| \phi_3(\phi_1(G(y))) - \phi_3(\phi_1(x)) \right \| \leqslant \gamma
\end{equation}
where $\gamma$ represents the range.
Accordingly, the objective function for $||\phi_3(\phi_1(x))-y||$ can be expressed as:
\begin{equation}\label{key6}
\mathcal{L}_2 = E_{x \sim p_{x}}\left [  \left \| F(x) - y  \right \| \right ] 
+ E_{y \sim p_{y}}\left [ max (\left \| (F(G(y)) - F(x)  \right \| -\gamma, 0) \right ]
\end{equation}
where $F=\phi_3(\phi_1(\cdot))$.
Intuitively, $\gamma$ can be taken as a constant, \textit{e.g.}, 0.01. Further experiments have found that adaptively decreasing $\gamma$ in the training process contributes to performance improvement for camera localization. 
During the adversarial training between $G$ and $D$, $\phi_3(\phi_1(x))$ is approaching $y$. Therefore,  $\left \|  \phi_3(\phi_1(x)) -y \right \|$ is decreasing with  iterations. Consequently, we adaptively decrease $\gamma$  as well, and suppose $\gamma$ is proportional to $\left \|  \phi_3(\phi_1(x)) -y \right \|$:
\begin{equation}\label{key7}
\gamma =k \cdot \left \|  \phi_3(\phi_1(x)) -y \right \|
\end{equation}
where $k$ is taken as 0.1 in the experiments. Experiments in Section \ref{Exp} have demonstrated that adaptively decreasing $\gamma$ is of better performance than taking $\gamma$ as a constant.

The term $\left \| \phi_3(\phi_1(G(y))) - \phi_3(\phi_1(x)) \right \| \leqslant \gamma$  can be regarded as the pose estimation of generated samples.   Pose estimation of fake examples can help roughly confirm the pose range of generated samples, further serving as a stronger constraint to help the conditional discrimination. We will see in Section \ref{Exp} that, pose estimation indeed contributes to performance improvement.

\subsection{Objective Function}
To sum up, we can obtain the objective function $\mathcal{L}_D$ for $D$:
\begin{equation}\label{key8}
\mathcal{L}_D = \mathcal{L}_1 + \alpha \cdot \mathcal{L}_2
\end{equation}
where $\mathcal{L}_1$ and $\mathcal{L}_2$ are shown in Equation (\ref{key5}) and (\ref{key6}), respectively, $\alpha$ is a hyperparameter and taken as 0.1 in the experiments.

For $G$, the objective function $\mathcal{L}_G$ can be expressed as:
\begin{equation}\label{key9}
\begin{split}
\mathcal{L}_G =& -E_{y \sim p_{y}}\left [ \varphi(\phi_1(G(y)))   \right ]\\
&+ \beta_1 \cdot E_{y \sim p_{y}}\left [ \left \|F(G(y)) - y  \right \| \right ]\\
&+ \beta_2 \cdot E_{y \sim p_{y}}\left [ \left \| G(y)-x  \right \| \right ]
\end{split}
\end{equation}
where $-E_{y \sim p_{y}}\left [ \varphi(\phi_1(G(y)))\right ]$ forces $G$ to generate samples towards the real distribution. $E_{y \sim p_{y}}\left [ \left \|F(G(y)) - y  \right \| \right ]$ constrains generated samples  corresponding to the correct poses.  $E_{y \sim p_{y}}\left [ \left \| G(y)-x  \right \| \right ]$ requires generated samples to be identical to real samples in the pixel domain. $\beta_1$ and $\beta_2$ are hyperparameters, and are taken as 0.5 and 10, respectively in the experiments.


\section{Quantitative Analysis} \label{Exp}

\subsection{Experiment Setup}
We follow the practices in the literatures \cite{Kendall20152938, Brahmbhatt20182616, Sattler20193302}, and use Cambridge Landmarks \cite{Kendall20152938} and 7 Scenes \cite{H4} datasets  to evaluate the performance of PoseGANs on camera localization. The architecture of PoseGANs is shown in Figure \ref{fig:fig1}. 
The optimization settings follow SN-GANs \cite{Miyato2018}. To be specific, the batch size is taken as 16, the learning rate is taken as 0.0002, the number of updates of the discriminator per one update of the generator $n_{critic}$ is 1, and Adam optimizer \cite{Kingma2014} is used as the optimization with the first and second order momentum parameters as 0 and 0.9, respectively. In addition, spectral normalization \cite{Miyato2018} is applied in $G$ and $D$ to guarantee the Lipschitz continuity, which contributes to stable training.
Experiments have been conducted based on samples of multiply resolutions, including 32$\times$32 pixels, 64$\times$64 pixels, 128$\times$128 pixels, 256$\times$256 pixels. PoseGANs have similar performances under multiply resolutions but have a increasing computational demands with the raising of image sizes. To present the synthesized images in a better way and consider the  computational efficiency, we conduct extensive experiments across different datasets with sample resolution of 128 $\times$ 128 pixels. 
Image augmentation is applied in the preprocessing. Specifically, the camera shot images in the datasets are compressed to 144 $\times$ 144 pixels, then random/center cropped to  128 $\times$ 128 pixels for training/evaluation.   A single Nvidia GTX1080ti is sufficient for PoseGANs, where 100k iterations takes about 12 hours for training, and less than 1 ms is used for computing the pose in the evaluations.

\begin{table*}[t]
	\centering
	\caption{Camera localization results on the Cambridge Landmarks and 7 Scenes dataset. We compare PoseGANs with other camera localization methods, and report the median position/orientation errors in meter/degree. LR is learning-based regression, IR is image retrieval and SL represents structure-based localization. ConfigA represents that only the first term in Equation \ref{key6} is used as the objective function for $D$, \textit{i..e}, no adversarial loss is applied. ConfigB means no pose estimation for generated images is used, \textit{i.e.}, the second term in Equation \ref{key6} is removed, and configC means pose estimation is applied but $\gamma$ in Equation \ref{key6} is taken as a constant 0.01. }
	\resizebox{\textwidth}{22mm}{
		\begin{tabular}{l |l| c| c| c| c||c| c| c| c| c| c|c}
			\hline \hline
			& & \multicolumn{4}{c||}{Cambridge Landmarks} &\multicolumn{7}{c}{7 Scenes}\\
			\cline{3-13}
			& &Kings&Old&Shop&St.Mary's&Chess&Fire&Heads&Office&Pumpkin&Kitchen&Stairs\\ 
			\hline
			&PoseGANs  & 1.22/4.44&1.52/4.91&\textbf{0.88/4.80}&1.82/5.79 &0.09/4.58&\textbf{0.24/9.46}&0.17/13.38&0.19/8.80&\textbf{0.16/6.28}&0.26/8.23&0.28/10.14\\
			&PoseGAN-configA&1.97/5.20&2.27/5.17&1.53/7.41&2.67/7.89&0.21/6.23&0.39/12.30&0.27/13.21&0.35/7.11&0.33/7.16&0.49/8.59&0.46/11.22\\
			&PoseGAN-configB&1.51/6.23&1.62/5.43&1.12/5.76&1.94/6.84&0.13/5.15&0.31/10.23&0.23/14.13&0.25/9.78&0.23/7.68&0.41/9.92&0.39/11.21\\
			&PoseGAN-configC&1.34/5.10&1.55/4.92&0.97/5.12&1.90/5.99&0.11/4.15&0.29/9.19&0.19/12.10&0.23/9.23&0.20/6.68&0.33/8.72&0.34/10.21\\
		
			\hline
			\multirow{8}{*}{\rotatebox{90}{LR}}&PoseNet (PN) \cite{Kendall20152938}&1.92/5.40& 2.31/5.38& 1.46/8.08& 2.65/8.48&0.32/8.12 &0.47/14.4& 0.29/12.0& 0.48/7.68 &0.47/8.42 &0.59/8.64 &0.47/13.8\\
			&PN learned weights \cite{Kendall20175974}&0.99/1.06& 2.17/2.94& 1.05/3.97& 1.49/3.43&0.14/4.50& 0.27/11.8& 0.18/12.1 &0.20/5.77 &0.25/4.82 &0.24/5.52 &0.37/10.6\\
			&Bay.PN \cite{Kendall20164762} &1.74/4.06& 2.57/5.14 &1.25/7.54 &2.11/8.38& 0.37/7.24& 0.43/13.7& 0.31/12.0& 0.48/8.04 &0.61/7.08& 0.58/7.54 &0.48/13.1\\
			&LSTM PN \cite{Walc2017627}& \textbf{0.99/3.65}& \textbf{1.51/4.29}& 1.18/7.44& \textbf{1.52/6.68}& 0.24/5.77& 0.34/11.9& 0.21/13.7& 0.30/8.08& 0.33/7.00& 0.37/8.83& 0.40/13.7\\
			&GPoseNet \cite{Cai20188} & 1.61/2.29& 2.62/3.89 &1.14/5.73 &2.93/6.46 &0.20/7.11& 0.38/12.3& 0.21/13.8& 0.28/ 8.83 &0.37/6.94& 0.35/8.15& 0.37/12.5\\
			&MapNet \cite{Brahmbhatt20182616} & 1.07/1.89 & 1.94/3.91  &1.49/4.22 & 2.00/4.53 & \textbf{0.08/3.25}  &0.27/11.7 & 0.18/13.3  &\textbf{0.17/5.15} & 0.22/4.02 & \textbf{0.23/4.93} & 0.30/12.1\\
			\hline
			\multirow{2}{*}{\rotatebox{90}{IR}}&DenseVLAD \cite{A15}& 2.80/5.72& 4.01/7.13 &1.11/7.61 &2.31/8.00&  0.21/12.5 &0.33/13.8 &0.15/14.9& 0.28/11.2& 0.31/11.3 &0.30/12.3& 0.25/15.8\\
			&DenseVLAD + Inter& 1.48/4.45& 2.68/4.63 &0.90/4.32 &1.62/6.06& 0.18/10.0& 0.33/12.4& \textbf{0.14/14.3}& 0.25/10.1& 0.26/9.42& 0.27/11.1& \textbf{0.24/14.7}\\
			\hline
			\multirow{2}{*}{\rotatebox{90}{SL}}&Active Search \cite{Sattler20161744}& 0.42/0.55& 0.44/1.01 &0.12/0.40& 0.19/0.54& 0.04/1.96& 0.03/1.53& 0.02/1.45& 0.09/3.61& 0.08/3.10 &0.07/3.37 &0.03/2.22\\
			&DSAC++ \cite{Brachmann20184654}&0.18/0.30& 0.20/0.30& 0.06/0.30& 0.13/0.40&0.02/0.50& 0.02/0.90& 0.01/0.80& 0.03/0.70& 0.04/1.10& 0.04/1.10& 0.09/2.60\\
			\hline \hline
	\end{tabular}}
	\label{t22}
\end{table*}

\subsection{Camera Localization Results}
To test the validity of proposed PoseGANs, comparative trials are conducted on Cambridge Landmarks and 7 scenes datasets, and results are listed in Table \ref{t22}.  ConfigA in Table \ref{t22} only applies the first term in Equation \ref{key6} as the objective function for $D$, \textit{i.e.}, $\mathcal{L}_D = E_{x \sim p_{x}}\left [  \left \| F(x) - y  \right \| \right ]$, and no adversarial loss is used in PoseGAN-configA. Then PoseGAN-configA turns to a regression model, resulting in about 75\% increase in pose estimation errors, which further demonstrates that adversarial training proposed by PoseGANs contributes to pose estimations.
ConfigB in Table \ref{t22} indicates that pose estimation of the generated samples $G(y)$ provides a stronger constraint to help conditional discrimination, thus contributing to about 20\% performance improvement. Besides, as indicated by configC, adaptively decreasing $\gamma$ performs better than taking it as a constant.

In addition, we compare the results with the benchmark approaches in Table \ref{t22}. PoseGANs and learning-based regressions are both CNNs based approches to the camera localization problem.
As clearly shown in Figure \ref{fig:fig1}, $\phi_3(\phi_1(\cdot))$ only contains 10 convolutional layers, about 0.1 $\times$ size of ResNet-101 based regression model, and 0.3 $\times$ size of ResNet-34 based regression model \cite{Kendall20152938}. Despite using a significantly simpler network, PoseGANs do not compromise on camera localization performance, instead, they achieve comparable results with learning-based regression models like PoseNet.  
We reason that the performance improvements may benefit from the generated samples provided by $G$. $D$ is trained on the supervision of the positive samples $(x, y)$, as well as the negative samples $(G(y), y)$, in contrast to learning-based regression models like PoseNet, where only positive samples are used.
To gain a deeper understanding of PoseGANs, we compare PoseGAN with other camera localization method in details.

\begin{figure*}[tp]
	\begin{subfigure}{1\textwidth}
		\centering
		\includegraphics[height=0.0975\linewidth,width=0.13\linewidth]{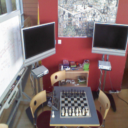} 
		\includegraphics[height=0.0975\linewidth,width=0.13\linewidth]{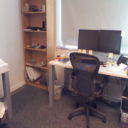}
		\includegraphics[height=0.0975\linewidth,width=0.13\linewidth]{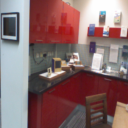}
		\includegraphics[height=0.0975\linewidth,width=0.13\linewidth]{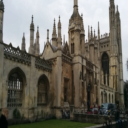}
		\includegraphics[height=0.0975\linewidth,width=0.13\linewidth]{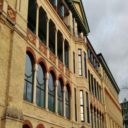}
		\includegraphics[height=0.0975\linewidth,width=0.13\linewidth]{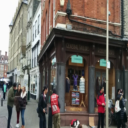}
		\includegraphics[height=0.0975\linewidth,width=0.13\linewidth]{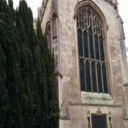}
		\caption{camera shot images.}
	\end{subfigure}	
	\newline
	\begin{subfigure}{1\textwidth}
		\centering
		\includegraphics[height=0.0975\linewidth,width=0.13\linewidth]{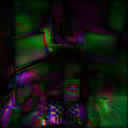}  
		\includegraphics[height=0.0975\linewidth,width=0.13\linewidth]{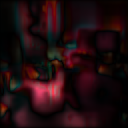}
		\includegraphics[height=0.0975\linewidth,width=0.13\linewidth]{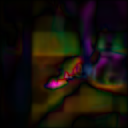}
		\includegraphics[height=0.0975\linewidth,width=0.13\linewidth]{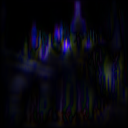}
		\includegraphics[height=0.0975\linewidth,width=0.13\linewidth]{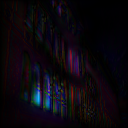}
		\includegraphics[height=0.0975\linewidth,width=0.13\linewidth]{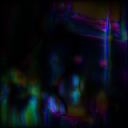}
		\includegraphics[height=0.0975\linewidth,width=0.13\linewidth]{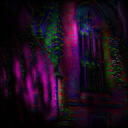}
		\caption{ Saliency map produced by PoseNet.}
	\end{subfigure}	
	\begin{subfigure}{1\textwidth}
		\centering
		\includegraphics[height=0.0975\linewidth,width=0.13\linewidth]{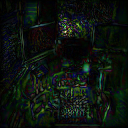}  
		\includegraphics[height=0.0975\linewidth,width=0.13\linewidth]{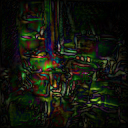}
		\includegraphics[height=0.0975\linewidth,width=0.13\linewidth]{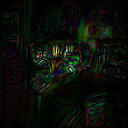}
		\includegraphics[height=0.0975\linewidth,width=0.13\linewidth]{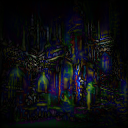}
		\includegraphics[height=0.0975\linewidth,width=0.13\linewidth]{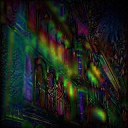}
		\includegraphics[height=0.0975\linewidth,width=0.13\linewidth]{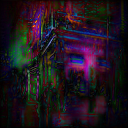}
		\includegraphics[height=0.0975\linewidth,width=0.13\linewidth]{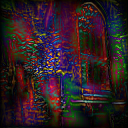}
		\caption{ Saliency map produced by PoseGANs.}
	\end{subfigure}	
	\begin{subfigure}{1\textwidth}
		\centering
		\includegraphics[height=0.0975\linewidth,width=0.13\linewidth]{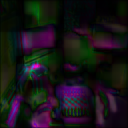}  
		\includegraphics[height=0.0975\linewidth,width=0.13\linewidth]{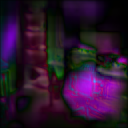}
		\includegraphics[height=0.0975\linewidth,width=0.13\linewidth]{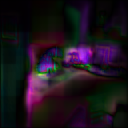}
		\includegraphics[height=0.0975\linewidth,width=0.13\linewidth]{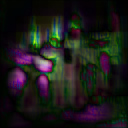}
		\includegraphics[height=0.0975\linewidth,width=0.13\linewidth]{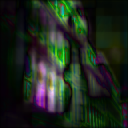}
		\includegraphics[height=0.0975\linewidth,width=0.13\linewidth]{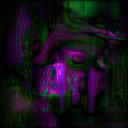}
		\includegraphics[height=0.0975\linewidth,width=0.13\linewidth]{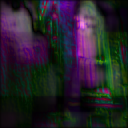}
		\caption{ Saliency map produced by PoseGAN-configA.}
	\end{subfigure}	
	\begin{subfigure}{1\textwidth}
		\centering
		\includegraphics[height=0.0975\linewidth,width=.13\linewidth]{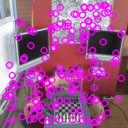}  
		\includegraphics[height=0.0975\linewidth,width=.13\linewidth]{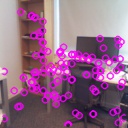}  
		\includegraphics[height=0.0975\linewidth,width=.13\linewidth]{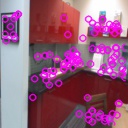}  
		\includegraphics[height=0.0975\linewidth,width=.13\linewidth]{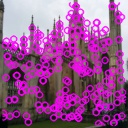}  
		\includegraphics[height=0.0975\linewidth,width=.13\linewidth]{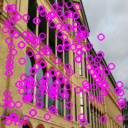}  
		\includegraphics[height=0.0975\linewidth,width=.13\linewidth]{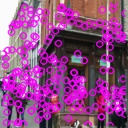}  
		\includegraphics[height=0.0975\linewidth,width=.13\linewidth]{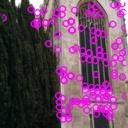}  
		\caption{Keypoints extracted by SIFT.}
	\end{subfigure}
	\begin{subfigure}{1\textwidth}
		\centering
		\includegraphics[height=0.0975\linewidth,width=0.13\linewidth]{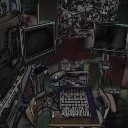}  
		\includegraphics[height=0.0975\linewidth,width=0.13\linewidth]{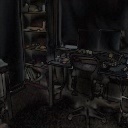}
		\includegraphics[height=0.0975\linewidth,width=0.13\linewidth]{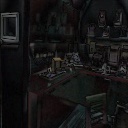}
		\includegraphics[height=0.0975\linewidth,width=0.13\linewidth]{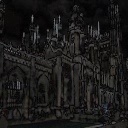}
		\includegraphics[height=0.0975\linewidth,width=0.13\linewidth]{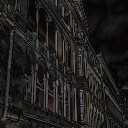}
		\includegraphics[height=0.0975\linewidth,width=0.13\linewidth]{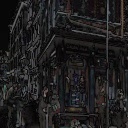}
		\includegraphics[height=0.0975\linewidth,width=0.13\linewidth]{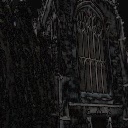}
		\caption{Structure component extracted by structure-texture decomposition algorithm. }
	\end{subfigure}	
	
	\caption{Illustration of the differences between PoseGANs and other camera localization methods. For the limitation of page lengths, we show  samples taken from the Chess, Office, Redlkitchen datasets (from 7 scenes) and the KingsCollege, OldHospital, ShopFacade, StMarysChurch datasets (from Cambridge Landmarks) in (a). (b), (c) and (d) visualize the saliency map produced by PoseNet, PoseGANs and PoseGAN-configA, respectively, where samples in (a) are used. (e) shows keypoints extracted by SIFT, and (f) shows structure components of the camera shot images.}
	\label{fig:fig2}
\end{figure*}

\subsection{Comparison with other Methods} \label{discussion}

PoseGANs and learning-based regression models rely on CNNs to achieve the goal of camera localization.
For this reason, the implementation of PoseGANs (Figure \ref{fig:fig1}) and the loss function (Equation (\ref{key8}))  may be erroneously identified as being identical to learning-based regression model. Actually, PoseGANs are fundamentally different from other camera localization methods including learning-based regressions.

All the camera localization algorithms rely on features to  conduct pose estimation. The saliency map \cite{Zeiler2014, Simonyan2013} calculates the magnitude of the gradient of the loss function with respect to the pixel intensities, and use the sensitivity of the pose with respect to the pixels as an indicator of how important the models considers different parts of the image. Thus, we show saliency maps produced by PoseNet \cite{Kendall20152938} and PoseGAN in Figure \ref{fig:fig2} (b) and (c), respectively to indicate the difference between PoseGAN and learning-based regression models, where PoseNet acts as a representation of learning-based regression models. The saliency maps clearly indicate how input images affect pose estimation results, and what features PoseNet and PoseGANs use to determine the pose. As a comparison, local features extracted by SIFT \cite{Lowe2004} are also plot in Figure \ref{fig:fig2} (e) to show what features traditional structure-based localization and some image retrieval algorithms use to accomplish the camera localization task. 

Firstly, it is clearly seen that PoseNet is more likely to use the local texture features in the pose estimation process, \textit{e.g.}, in the Chess scene, PoseNet mostly uses local texture features like the chessboard and the TV (the corresponding parts in the saliency map is highlighted). Similar phenomena is found in all other experiments conducted on different sequences.
In contrast, PoseGANs pay more attention to the geometry structures in the decision process, \textit{i.e.}, the highlighted parts in Figure \ref{fig:fig2} (c) roughly outline the structure information in the scene. 
To better understand the geometry structure, we use a structure-texture decomposition algorithm \cite{H1} to extract the structure component of each camera shot image, and show them in Figure \ref{fig:fig2} (f). We can see that, the highlighted parts in the saliency map for PoseGANs (Figure \ref{fig:fig6} (c)) are consistent with the structure components in Figure \ref{fig:fig2} (f), indicating that PoseGANs are geometric structure motivated.

Figure \ref{fig:fig2}(d) shows the saliency map produced by PoseGAN-configA. It is clearly seen that PoseGAN-configA mainly focus on local texture features instead of geometry structure, different from PoseGANs but similar to PoseNet. Removing the adversarial loss in Equation \ref{key8} leads to the PoseGAN-configA incapable of being sensitive to the geometry structure, indicating that it is due to the adversarial loss in Equation \ref{key8} that PoseGANs exploit the geometric structures.

\begin{figure*}[htp]	
	\begin{subfigure}{.155\textwidth}
		\centering 
		\includegraphics[height=0.75\linewidth,width=1\linewidth]{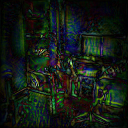}
		\caption{}
	\end{subfigure}	
	\begin{subfigure}{.155\textwidth}
		\centering 
		\includegraphics[height=0.75\linewidth,width=1\linewidth]{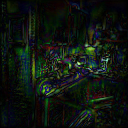}
		\caption{}
	\end{subfigure}	
	\begin{subfigure}{.155\textwidth}
		\centering 
		\includegraphics[height=0.75\linewidth,width=1\linewidth]{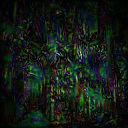}
		\caption{}
	\end{subfigure}	
	\begin{subfigure}{.155\textwidth}
		\centering 
		\includegraphics[height=0.75\linewidth,width=1\linewidth]{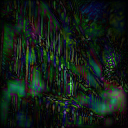}
		\caption{}
	\end{subfigure}	
	\begin{subfigure}{.155\textwidth}
		\centering 
		\includegraphics[height=0.75\linewidth,width=1\linewidth]{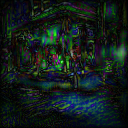}
		\caption{}
	\end{subfigure}	
	\begin{subfigure}{.155\textwidth}
		\centering 
		\includegraphics[height=0.75\linewidth,width=1\linewidth]{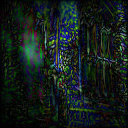}
		\caption{}
	\end{subfigure}	
	\caption{Saliency maps produced by PoseGANs, where training is based on the Chess dataset, while visualization is based on samples from the (a) Office, (b) Redlkitchen, (c) KingsCollege, (d) OldHospital, (e) ShopFacade and (f) StMarysChurch datasets, which correspond to the samples in Figure \ref{fig:fig2} (a).}
	\label{fig:fig3}
	
\end{figure*}

Furthermore, we discover that the structure awareness is not for specific data  but a property of PoseGANs.
To be specific, we train PoseGAN on the Chess dataset, then visualize the saliency maps using samples from other datasets.
The results are illustrated in Figure \ref{fig:fig3}. Although PoseGANs are trained using samples from the Chess dataset, they are capable of exploiting the geometry structures of not only samples from the Chess dataset, but also samples from other datasets, demonstrating that PoseGANs have been equipped with structure-aware ability, which is available for various scenes. 

\begin{figure}[htp]
	\begin{subfigure}{0.32\textwidth}
		\centering
		\includegraphics[height=0.5\linewidth, width=0.66\linewidth]{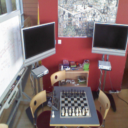}\\
		\includegraphics[height=0.5\linewidth, width=0.66\linewidth]{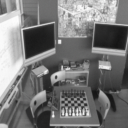}
		\caption{}
	\end{subfigure}
	\begin{subfigure}{0.32\textwidth}
		\centering
		\includegraphics[height=0.5\linewidth, width=0.66\linewidth]{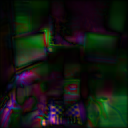}\\
		\includegraphics[height=0.5\linewidth, width=0.66\linewidth]{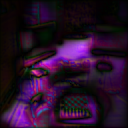}
		\caption{}
	\end{subfigure}
	\begin{subfigure}{0.32\textwidth}
		\centering
		\includegraphics[height=0.5\linewidth, width=0.66\linewidth]{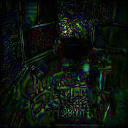}\\
		\includegraphics[height=0.5\linewidth, width=0.66\linewidth]{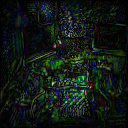}
		\caption{}
	\end{subfigure}
	\vspace*{-0.4cm}
	\caption{Illustration of the superior of geometry structure over local texture features. We turn the camera shot image to grayscale, and show an example from the Chess dataset in (a). (b) and (c) plot the saliency maps produced by PoseNet and PoseGAN, respectively, where the sample in (a) is used.}
	\label{fig:fig4}
	
\end{figure}

It is obvious that geometry structure is more reliable than local texture features. To elaborate this, we turn the camera shot image to grayscale, then visualize the saliency maps for PoeseNet and PoseGANs in Figure \ref{fig:fig4}. It is clearly seen that the highlighted parts for PoseNet (Figure \ref{fig:fig4} (b)) have changed significantly. As a comparison, the saliency map for PoseGANs is hardly affected. In addition, we  turn samples in the test sets to grayscale mode, then conduct pose estimation. Results are listed in Table \ref{t23}. It is clearly seen that PoseGANs are less affected by the grayscale images than PoseNet, indicating that geometry structure is more reliable in the pose estimation process over local texture features.

\begin{table*}[ht]
	\centering
	\caption{Camera localization results under grayscale mode. We turn the camera shot images in the test sets to grayscale, then report the median position/orientation errors in meter/degree. $\Delta$ means the average increase in position/orientation estimation errors when compared with the results in Table \ref{t22}.}
	\resizebox{\textwidth}{6mm}{
		\begin{tabular}{c| c| c| c| c| c||c| c| c| c| c| c|c|c}
			\hline \hline
			& \multicolumn{5}{c||}{Cambridge Landmarks} &\multicolumn{8}{c}{7 Scenes}\\
			\cline{2-14}
			& Kings&Old&Shop&St.Mary's&$\Delta$&Chess&Fire&Heads&Office&Pumpkin&Kitchen&Stairs&$\Delta$\\ 
			\hline	
			PoseGANs& 8.59/14.20&7.83/25.26&3.66/13.84&10.48/8.51&6.28$\uparrow$/10.46$\uparrow$ &0.35/12.03&0.47/25.87&0.32/10.79&0.29/11.89&0.48/28.63&0.85/21.10&0.28/10.14&0.21$\uparrow$/7.92$\uparrow$\\
			\hline
			PoseNet&9.68/16.25& 14.65/30.83& 5.11/17.51& 23.99/25.87&11.27$\uparrow$/15.78$\uparrow$
			&0.74/16.24 &0.65/20.51& 0.52/16.71& 0.88/12.68 &0.74/20.12 &1.15/29.39 &0.65/18.22&0.32$\uparrow$/8.68$\uparrow$\\
			\hline
			
	\end{tabular}}
	\label{t23}
\end{table*}

\begin{figure}[t]
	\centering
	\includegraphics[height=0.25\linewidth]{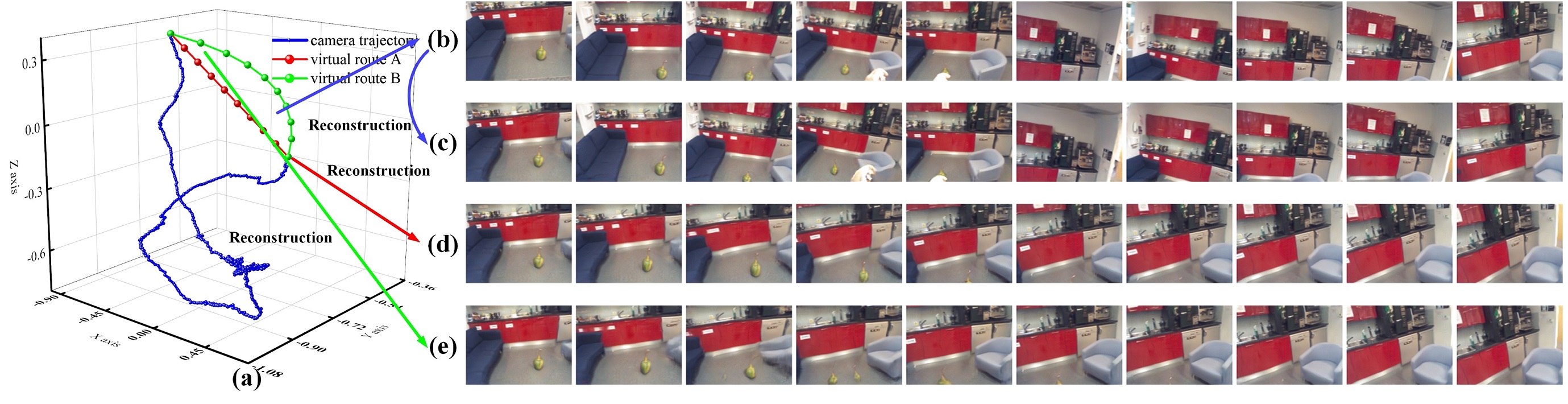}	
	\caption{View synthesis based on the Pumpkin dataset by the proposed PoseGANs. (a) camera trajectory of seq-01 in the Pumpkin dataset. (b) corresponding camera shot images along the camera's trajectory. (c) generated images along the camera's trajectory. (d) and (e) are images synthesized along  virtual routes A and B where the poses along the routes are obtained by linear and parabola interpolations between the start and the end points of the camera's trajectory, respectively.}
	\label{fig:fig5}
\end{figure}

Secondly, PoseGANs are able to establish the correspondence between the 2D images and the scene, which lead us to reason that PoseGANs may be aware of projective geometry.
For structure-based localization, features as shown in Figure \ref{fig:fig2} (e)  are used to establish correspondence between 2D pixel positions and 3D points coordinates in the scene (2D-3D matches), where knowledge of projection geometry is utilized.
Unlike the 2D-3D matches in traditional structure-based localization but in a similar way, PoseGANs establish the correspondence between the 2D images and scenes, \textit{i.e.},  given the pose, the generator $G$ allows PoseGANs to generate its corresponding camera shot images, which is referred to as view synthesis.
Figure \ref{fig:fig5} gives an example of view synthesis in PoseGANs. 
In addition to generating images along the real camera trajectory, PoseGANs can generate images along virtual routes. Figure \ref{fig:fig5} (d) and (e) plot synthesized images  along virtual routes A and B where the poses along the routes are obtained by linear and parabola interpolations between the start and the end points of the camera’s trajectory, respectively. Even though some points in the virtual routes are far from the real camera trajectory, synthesized images are of high quality and accord well with the scene, demonstrating that PoseGANs have established the correspondence between the 2D images and the  scene.

Knowledge of projective geometry is essential to establish the correspondence between the 2D images and 3D scene in the traditional structure-based localization. From this perspective, the view synthesis technique demonstrates that PoseGANs may be aware of projective geometry in the pose-to-image translation.
However, learning-based regression algorithms, also relying on CNNs to conduct camera localization, do not  use knowledge about projective geometry. Rather, they learn the mapping from image content to camera pose from data \cite{Sattler20193302}, further demonstrating the superior of PoseGANs over learning-based regressions.

To summarize, we can see that PoseGANs tackle the camera localization problem via a pose-to-image translation, therefore differ in principle from other camera localization methods. Secondly, PoseGANs are aware of geometry structure in the scene, and exploit the geometry structures in the scene to estimate the camera pose instead of the local texture features. Thirdly, PoseGANs establish the correspondence between 2D images and scenes in a implicit way, \textit{i.e}, given the pose, PoseGANs are able to obtain its corresponding images.

\section{Extended Applications}
In addition to camera localization and view synthesis, PoseGANs are capable of applications like moving object elimination and frame interpolation.

\subsection{Moving Object Elimination}
\begin{figure*}[tp]
	\begin{subfigure}{0.5\textwidth}
		\centering
		\includegraphics[height=0.15\linewidth, width=1\linewidth]{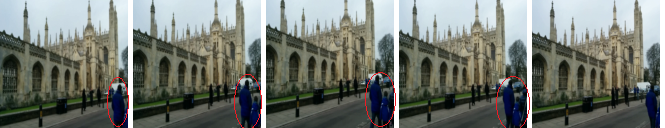}  \\
		\includegraphics[height=0.15\linewidth, width=1\linewidth]{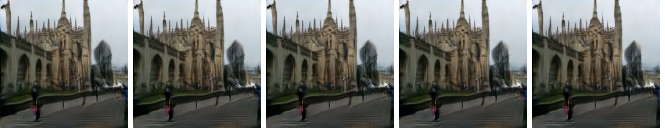}  
		\vspace*{-0.5cm}
		\caption{KingsCollege}
	\end{subfigure}
	\begin{subfigure}{0.5\textwidth}
		\centering
		\includegraphics[height=0.15\linewidth, width=1\linewidth]{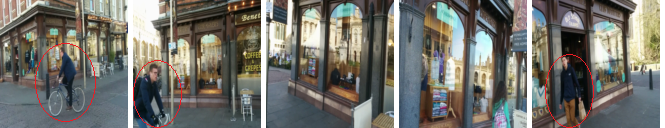}  \\
		\includegraphics[height=0.15\linewidth, width=1\linewidth]{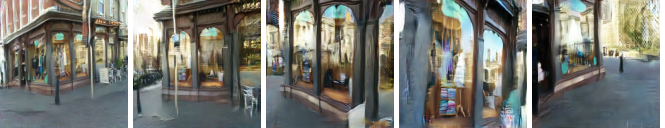}
		\vspace*{-0.5cm}
		\caption{ShopFacade}
	\end{subfigure}
	\vspace*{-0.4cm}
	\caption{Moving Object Elimination. (a) seq-01 of the KingsCollege dataset, (b) seq-02 of the ShopFacade dataset. Original images from datasets are at the top, and generated images are at the bottom. Red circles in (a) and (b) mark pedestrians. As a comparison, PoseGANs only synthesize the scene, and remove the pedestrians in the scene. }
	\label{fig:fig6}
	\vspace*{-0.0cm}
	
	\begin{subfigure}[b]{0.09\textwidth}
		\centering
		\includegraphics[height=0.75\linewidth, width=1\linewidth]{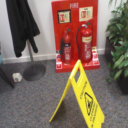}  
		\caption{}
	\end{subfigure}	
	\begin{subfigure}[b]{0.757\textwidth}
		\centering
		\includegraphics[height=0.09\linewidth, width=1\linewidth]{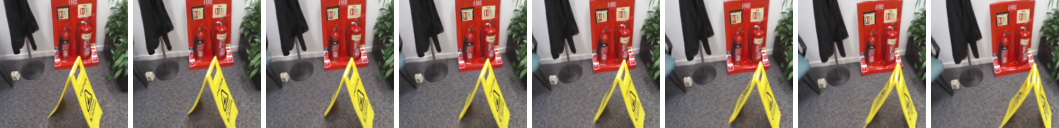}  
		\caption{}
	\end{subfigure}	
	\begin{subfigure}[b]{0.09\textwidth}
		\centering
		\includegraphics[height=0.75\linewidth, width=1\linewidth]{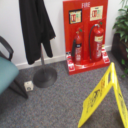}  
		\caption{}
	\end{subfigure}	
	
	\caption{Frame Interpolation of seq-01 in the Fire dataset. (a) and (c) are the start and end frame, respectively, which are taken from the Fire dataset, and (b) shows frames interpolated between (a) and (c). }	
	\label{fig:fig7}
	\vspace*{-0.5cm}	
	
\end{figure*}

$p_x$ is uesd to describe the latent distribution the camera shot image obeys. Obviously, $p_x$ describes the relative geometry relationship among all the static objects in the scene.  
The geometry relationship between the moving objects and the static objects is varying, thus can hardly be catched by $p_x$. For this reason, the view synthesis in PoseGAN would be free from the influence of the moving objects.
Figure \ref{fig:fig6} shows the view synthesis based on the KingsColleage and ShopFacade datasets, where the original camera shot images are affected by pedestrians. It is clearly seen that pedestrians, marked in red circles, appear in the original scenes, have disappeared in the synthesized scenes, which is exactly what we have expected. 
In addition, the scene occluded by pedestrians is also well-reconstructed, demonstrating that PoseGANs are capable of moving object elimination.

\subsection{Frame Interpolation}

The ability of view synthesis at a given pose allows PoseGAN to perform frame interpolation, which will have the effect of increasing frame rate.
In Figure \ref{fig:fig7}, we show frames generated between two given frames. We obtain the pose of the $i$-th generated frame $y_i$ by linearly interpolating between the starting frame's pose $y_s$ and the ending frame's pose $y_e$, and then use the poses to generate corresponding frames.
It is seen in Figure \ref{fig:fig7} that the scene transits smoothly from the start to the end frames, demonstrating the ability of PoseGANs in creating realistic sequences with higher frame rates. 
In practice, we can insert an arbitrary number of frames between any given two frames.
In the accompanying videos, we show examples of inserting 10 frames between two consecutive frames, effectively increasing frame rate 10 times.
It is seen that the interpolated videos show much smoother transitions between frames in contrast to the jerky appearances of the original sequences.

\section{Concluding Remarks}
\subsection{Conclusions}
In this paper, we introduce PoseGANs to solve camera localization via a pose-to-image translation. To implement PoseGANs, we design the architecture and introduce the loss objectives. 
We experimentally confirm the effectiveness of the proposed architecture and loss objectives. Besides, PoseGANs are demonstrated to have good performance on camera localization, and we provide examples to explain the differences between PoseGANs and other methods for camera localization. In addition to camera localization, PoseGANs are capable of a few application,  \textit{e.g.}, view synthesis, moving object elimination and frame interpolation.

\subsection{Discussion}
In the training of PoseGANs, images in the training sets are regarded as the positive samples, and the generated samples, whose poses are identical to that of the corresponding camera shot images, are treated as negative ones. The number of the positive samples is equal to that of negative ones.
As shown in Figure \ref{fig:fig5}, PoseGANs are capable of generating images, whose poses are unavailable in the datasets. However, those generated images at new poses are not utilized in PoseGANs. Because introducing those images would lead to the unbalance between positive and negative samples, further leading to the training instability problem. Worst of all, PoseGAN would fail to synthesize images and have a bad performance on predicting the poses. Thus, efforts can be made on how to utilize those synthesized images in PoseGANs, which may potentially improve the performance. Besides, those generated samples may be used in traditional SfM technique or learning-based regression models as an extension to the original datasets.

In addition to camera localization, the pose-to-image translation framework may be used in other 3D vision tasks like depth estimation. Furthermore, the depth channel may be also used in PoseGANs to help improve the quality of generated images. Overall, PoseGANs have a number of potential applications. 

\section*{Acknowledgement}
This work is partially supported by the Education Department of Guangdong Province, P.R. China, under project No.2019KZDZX1028.

\section*{References}

\bibliography{posegan}

\begin{thebibliography}{10}
\expandafter\ifx\csname url\endcsname\relax
  \def\url#1{\texttt{#1}}\fi
\expandafter\ifx\csname urlprefix\endcsname\relax\def\urlprefix{URL }\fi
\expandafter\ifx\csname href\endcsname\relax
  \def\href#1#2{#2} \def\path#1{#1}\fi

\bibitem{Engel2014834}
J.~Engel, T.~Schöps, D.~Cremers, Lsd-slam: Large-scale direct monocular slam,
  European conference on computer vision (2014) 834--849.

\bibitem{MurArtal20151147}
R.~MurArtal, J.~Montiel, J.~D. Tardos, Orbslam: a versatile and accurate
  monocular slam system, IEEE transactions on robotics 31~(5) (2015)
  1147--1163.

\bibitem{Zhou20151364}
H.~Zhou, D.~Zou, L.~Pei, R.~Ying, P.~Liu, W.~Yu, Structslam: Visual slam with
  building structure lines, IEEE Transactions on Vehicular Technology 64~(4)
  (2015) 1364--1375.

\bibitem{Hane201714}
C.~Häne, L.~Heng, G.~H. Lee, F.~Fraundorfer, P.~Furgale, T.~Sattler,
  M.~Pollefeys, 3d visual perception for self-driving cars using a multi-camera
  system: Calibration, mapping, localization, and obstacle detection, Image and
  Vision Computing 68 (2017) 14--27.

\bibitem{Castle200815}
R.~Castle, G.~Klein, D.~W. Murray, Video-rate localization in multiple maps for
  wearable augmented reality, 2008 12th IEEE International Symposium on
  Wearable Computers (2008) 15--22.

\bibitem{Brachmann20176684}
E.~Brachmann, A.~Krull, S.~Nowozin, J.~Shotton, F.~Michel, S.~Gumhold,
  C.~Rother, Dsac-differentiable ransac for camera localization, In Proceedings
  of the IEEE Conference on Computer Vision and Pattern Recognition (2017)
  6684--6692.

\bibitem{Brachmann20184654}
E.~Brachmann, C.~Rothe, Learning less is more - 6d camera localization via 3d
  surface regression, In Proceedings of the IEEE Conference on Computer Vision
  and Pattern Recognition (2018) 4654--4662.

\bibitem{Cavallari20174457}
T.~Cavallari, S.~Golodetz, N.~A. Lord, J.~Valentin, L.~D. Stefano, P.~H. Torr,
  On-the-fly adaptation of regression forests for online camera relocalisation,
  In Proceedings of the IEEE Conference on Computer Vision and Pattern
  Recognition (2017) 4457--4466.

\bibitem{Meng20176886}
L.~Meng, J.~Chen, F.~Tung, J.~J. Little, J.~Valentin, C.~W.~D. Silva,
  Backtracking regression forests for accurate camera relocalization, In 2017
  IEEE/RSJ International Conference on Intelligent Robots and Systems (IROS)
  (2017) 6886--6893.

\bibitem{Sattler20161744}
T.~Sattler, B.~Leibe, L.~Kobbelt, Efficient and effective prioritized matching
  for large-scale image-based localization, IEEE transactions on pattern
  analysis and machine intelligence 39~(9) (2016) 1744--1756.

\bibitem{Radwan20184407}
N.~Radwan, A.~Valada, W.~Burgard, Vlocnet++: Deep multitask learning for
  semantic visual localization and odometry, IEEE Robotics and Automation
  Letters 3~(4) (2018) 4407--4414.

\bibitem{Kendall20152938}
A.~Kendall, M.~Grimes, R.~Cipolla, Posenet: A convolutional network for
  real-time 6-dof camera relocalization, In Proceedings of the IEEE
  international conference on computer vision (2015) 2938--2946.

\bibitem{Melekhov2017879}
I.~Melekhov, J.~Ylioinas, J.~Kannala, E.~Rahtu, Image-based localization using
  hourglass networks, Proceedings of the IEEE International Conference on
  Computer Vision (2017) 879--886.

\bibitem{Kendall20164762}
A.~Kendall, R.~Cipolla, Modeling uncertainty in deep learning for camera
  relocalization, IEEE International Conference on Robotics and Automation
  (ICRA) (2016) 4762--4769.

\bibitem{Kendall20175974}
A.~Kendall, R.~Cipolla, Geometric loss functions for camera pose regression
  with deep learning, Proceedings of the IEEE Conference on Computer Vision and
  Pattern Recognition (2017) 5974--5983.

\bibitem{Brahmbhatt20182616}
S.~Brahmbhatt, J.~Gu, K.~Kim, J.~Hays, J.~Kautz, Geometry-aware learning of
  maps for camera localization, Proceedings of the IEEE Conference on Computer
  Vision and Pattern Recognition (2018) 2616--2625.

\bibitem{Ummenhofer20175038}
B.~Ummenhofer, H.~Zhou, J.~Uhrig, N.~Mayer, E.~Ilg, A.~Dosovitskiy, T.~Brox,
  Demon: Depth and motion network for learning monocular stereo, Proceedings of
  the IEEE Conference on Computer Vision and Pattern (2017) 5038--5047.

\bibitem{Albl20152292}
C.~Albl, Z.~Kukelova, T.~Pajdla, R6p-rolling shutter absolute camera pose, In
  Proceedings of the IEEE Conference on Computer Vision and Pattern Recognition
  (2015) 2292--2300.

\bibitem{Chum20081472}
O.~Chum, J.~Matas, Optimal randomized ransac, IEEE Transactions on Pattern
  Analysis and Machine Intelligence 30~(8) (2008) 1472--1482.

\bibitem{Fischler1981381}
M.~A. Fischler, R.~C. Bolles, Random sample consensus: a paradigm for model
  fitting with applications to image analysis and automated cartography,
  Communications of the ACM 24~(6) (1981) 381--395.

\bibitem{Laskar2017929}
Z.~Laskar, I.~Melekhov, S.~Kalia, J.~Kannala, Camera relocalization by
  computing pairwise relative poses using convolutional neural network,
  Proceedings of the IEEE International Conference on Computer Vision (2017)
  929--938.

\bibitem{Sattler20193302}
T.~Sattler, Q.~Zhou, M.~Pollefeys, L.~Leal-Taixe, Understanding the limitations
  of cnn-based absolute camera pose regression, In Proceedings of the IEEE
  Conference on Computer Vision and Pattern Recognition (2019) 3302--3312.

\bibitem{Simonyan2014}
K.~Simonyan, A.~Zisserman, Very deep convolutional networks for large-scale
  image recognition, arXiv preprint arXiv:1409.1556.

\bibitem{He2016}
K.~He, X.~Zhang, S.~Ren, J.~Sun, Deep residual learning for image recognition,
  Proceedings of the IEEE conference on computer vision and pattern
  recognition.

\bibitem{Balntas2018751}
V.~Balntas, S.~Li, V.~Prisacariu, Relocnet: Continuous metric learning
  relocalisation using neural nets, Proceedings of the European Conference on
  Computer Vision (ECCV) (2018) 751--767.

\bibitem{Melekhov2017675}
I.~Melekhov, J.~Ylioinas, J.~Kannala, E.~Rahtu, Relative camera pose estimation
  using convolutional neural networks, International Conference on Advanced
  Concepts for Intelligent Vision Systems (2017) 675--687.

\bibitem{Saha2018}
S.~Saha, G.~Varma, C.~V. Jawahar, Improved visual relocalization by discovering
  anchor points, arXiv1811.04370.

\bibitem{Goodfellow20142672}
I.~Goodfellow, J.~Pouget-Abadie, M.~Mirza, B.~Xu, D.~Warde-Farley, S.~Ozair,
  A.~Courville, Y.~Bengio, Generative adversarial nets, In Advances in neural
  information processing systems (2014) 2672--2680.

\bibitem{Radford2015}
A.~Radford, L.~Metz, S.~Chintala, Unsupervised representation learning with
  deep convolutional generative adversarial networks, arXiv preprint
  arXiv1511.06434.

\bibitem{Miyato2018}
T.~Miyato, T.~Kataoka, M.~Koyama, Spectral normalization for generative
  adversarial networks, arXiv preprint arXiv1802.05957.

\bibitem{Gulrajani2017}
I.~Gulrajani, F.~Ahmed, M.~Arjovsky, Improved training of wasserstein gans,
  Advances in Neural Information Processing Systems (2017) 5769--5779.

\bibitem{Arjovsky2017}
M.~Arjovsky, S.~Chintala, L.~Bottou, Wasserstein gan, arXiv preprint
  arXiv1701.07875.

\bibitem{Miyato20182}
T.~Miyato, M.~Koyama, cgans with projection discriminator, arXiv preprint
  arXiv1808.05637.

\bibitem{Andrew2018}
B.~Andrew, J.~Donahue, K.~Simonyan, Large scale gan training for high fidelity
  natural image synthesis, arXiv preprint arXiv1809.11096.

\bibitem{Karras2019}
T.~Karras, S.~Laine, T.~Aila, A style-based generator architecture for
  generative adversarial networks, In Proceedings of the IEEE Conference on
  Computer Vision and Pattern Recognition (2019) 4401--4410.

\bibitem{E15}
T.~Park, M.~Liu, T.~Wang, J.~Zhu, Semantic image synthesis with
  spatially-adaptive normalization, In Proceedings of the IEEE Conference on
  Computer Vision and Pattern Recognition (2019) 2337--2346.

\bibitem{Zhu2017}
J.~Zhu, T.~Park, P.~Isola, A.~A. Efros, Unpaired image-to-image translation
  using cycle-consistent adversarial networks, In Proceedings of the IEEE
  international conference on computer vision (2017) 2223--2232.

\bibitem{Ioffe2015}
S.~Ioffe, C.~Szegedy, Batch normalization: Accelerating deep network training
  by reducing internal covariate shift, arXiv preprint arXiv:1502.03167.

\bibitem{Glorot2011}
X.~Glorot, A.~Bordes, Y.~Bengio, Deep sparse rectifier neural networks,
  Proceedings of the fourteenth international conference on artificial
  intelligence and statistics.

\bibitem{H4}
B.~Glocker, S.~Izadi, J.~Shotton, A.~Criminisi, Real-time rgb-d camera
  relocalization, 2013 IEEE International Symposium on Mixed and Augmented
  Reality (ISMAR) (2013) 173--179.

\bibitem{Kingma2014}
D.~Kingma, J.~Ba, Adam: A method for stochastic optimization, arXiv preprint
  arXiv:1412.6980.

\bibitem{Walc2017627}
F.~Walch, C.~Hazirbas, L.~Leal-Taixe, T.~Sattler, S.~Hilsenbeck, D.~Cremers,
  Image-based localization using lstms for structured feature correlation, In
  Proceedings of the IEEE International Conference on Computer Vision (2017)
  627--637.

\bibitem{Cai20188}
M.~Cai, C.~Shen, I.~D. Reid, A hybrid probabilistic model for camera
  relocalization, BMVC 1~(2) (2018) 8.

\bibitem{A15}
A.~Torii, R.~Arandjelovic, J.~Sivic, M.~Okutomi, T.~Pajdla, 24/7 place
  recognition by view synthesis, In Proceedings of the IEEE Conference on
  Computer Vision and Pattern Recognition (2015) 1808--1817.

\bibitem{Zeiler2014}
M.~D. Zeiler, R.~Fergus, Visualizing and understanding convolutional networks,
  European conference on computer vision (2014) 818--833.

\bibitem{Simonyan2013}
K.~Simonyan, A.~Vedaldi, A.~Zisserman, Deep inside convolutional networks:
  Visualising image classification models and saliency maps, European
  conference on computer vision.

\bibitem{Lowe2004}
D.~Lowe, Distinctive image features from scale-invariant keypoints,
  International journal of computer vision 60~(2) (2004) 91--110.

\bibitem{H1}
L.~Xu, Q.~Yan, Y.Xia, J.~Jia, Structure extraction from texture via relative
  total variation, ACM Transactions on Graphics (TOG) 31~(6) (2012) 139.

\end{thebibliography}

\end{document}